# Detection of Distracted Driver using Convolution Neural Network

Narayana Darapaneni
Director - AIML
*Great Learning/Northwestern University*
*Illinois, USA*
darapaneni@gmail.com

Naman Vig
Student – *AIML*
*Great Learning*
*Gurgaon, India*
namanvig@gmail.com

Anwesh Reddy Paduri
Senior Data Scientist- AIML
*Great Learning*
*Pune, India*
anwesh@greatlearning.in

Jai Arora
Student – *AIML*
*Great Learning*
*Gurgaon, India*
jai.arora@gmail.com

Simrandeep Singh Gandhi
Student – *AIML*
*Great Learning*
*Gurgaon, India*
gandhi.simrandeepsingh@gmail.com

MoniShankar Hazra
Student – *AIML*
*Great Learning*
*Gurgaon, India*
monishankar@gmail.com

Saurabh Gupta
Student – *AIML*
*Great Learning*
*Gurgaon, India*
mail2saurabhgupta@gmail.com

*Abstract*— **With over 50 million car sales annually and over 1.3 million deaths every year due to motor accidents we have chosen this space. India accounts for 11 per cent of global death in road accidents. Drivers are held responsible for 78% of accidents. Road safety problems in developing countries is a major concern and human behavior is ascribed as one of the main causes and accelerators of road safety problems. Driver distraction has been identified as the main reason for accidents. Distractions can be caused due to reasons such as mobile usage, drinking, operating instruments, facial makeup, social interaction. For the scope of this project, we will focus on building a highly efficient ML model to classify different driver distractions at runtime using computer vision. We would also analyze the overall speed and scalability of the model in order to be able to set it up on an edge device. We use CNN, VGG-16, RestNet50 and ensemble of CNN to predict the classes.**

## I. Introduction

The problem at hand is related to the Automotive domain. With over 50 million car sales annually and over 1.3 million deaths every year due to motor accidents we have chosen this space. India accounts for 11 per cent of global death in road accidents [1]. In FY 18-19 motor insurance claims amounted to Rs. 58456.932 crores for India [2]. Each year, about three to five per cent of the country's GDP is invested in road accidents [3].

Drivers are held responsible for 78% of accidents [4]. Road safety problems in developing countries is a major concern and human behavior is ascribed as one of the main causes and accelerators of road safety problems [12]. With automotive becoming a tech-oriented domain due to the advent of driverless, fully connected vehicles, the current period allows viability of solving the problem. With AI a real-time alert system can be developed which can help in reminding the driver to stay focused, which in turn results in reducing the accidents bringing down the loss to property and life.

Driver distraction has been identified as the main reason for accidents. Distractions can be caused due to reasons such as mobile usage, drinking, operating instruments, facial makeup, social interaction. We propose to detect such distractions in real-time and alert the user to prevent any adversity. This solution can be deployed on an edge device setup on the vehicle to give an instant alarm and it can also interact through IoT to process data and give insights in asynchronous mode.

For the scope of this project, we will focus on building a highly efficient ML model to classify different driver distractions at runtime using computer vision. We would also analyze the overall speed and scalability of the model in order to be able to set it up on an edge device.

## II. Related Work

The driver distraction can be measured in visual, manual or cognitive ways. Under the manual and cognitive approach Aksjonov et al. [13] monitored lane maintenance and speed performance on specified road segments to detect driver's distraction. Saito et al. [14] used lane departure duration to develop an assistance system for predicting the driver's state. Apostoloff and Zelinsky [15] studied the driver's attention to lane maintenance task. Castignani et al. [16] by monitoring the acceleration, braking and steering activities of the driver, driving events were classified as risky or not. The system was named SenseFleet. A statistical analysis to determine the relation between driver distractions and the speed, acceleration, brake force, steering and lane position was presented by

Pavlidis et al. [17]. Wang et al. [18] proposed a forward collision warning algorithm that depends on the driver's braking activity.

Visual measurements, such as eye gaze, pupil diameter, head pose, facial expressions, driving posture etc can be used to detect driver distractions. Methods for detecting driver distraction can be grouped into three subgroups: mathematical models, rule-based models and models that are based on ML algorithms [19]. Scope of this study is limited to detecting driver distractions using models based on ML Algorithms. Baheti, B et al. [6] developed a support vector machine-based (SVM-based) model which extracts features of an image to detect the use of cell phones by drivers in a car. The dataset used images of the driver's face (frontal view) divided into two categories with phone and without. A SVM classifier was trained to detect the driver actions. Polynomial kernel (SVM) was the most advantageous kernel system with 91.57% accuracy among Linear, polynomial, RBF & Sigmoid. The study also used parallel processing to test real time video feed by simulating the videos on a computer.

A Convolutional Neural Network based system for distracted driver was developed for detection of different driver action [7]. The pre-trained ImageNet model was used for weight initialization and concept of transfer learning was applied. A modified VGG-16 was used which provided an accuracy of 96.31%. The original VGG-16 model overfitted the training set producing 100% accuracy. The author also suggested that due to similar postures the classes 'safe driving' and 'talking to passenger' were confused with each other. The model used an image dataset created by Abouelnaga et al. containing 10 classes and 17308 images [8]. Similarly, Alshalfan, Khalid & Zakariah, Mohammed [9] initialized the weights using the ImageNet model, followed by the application of the transfer learning concept. The modified VGG architecture achieved an accuracy of approximately 96.95%. They used the StateFarm dataset containing 9 classes and upward of 33,000 images. The model performed worst with same postures like 'talking with the right' and 'texting with the right' classes.

An Ensemble of Convolutional Neural Networks was also used for identification of driver distraction [10] It Consisted of a genetically weighted ensemble of convolutional neural networks. An AlexNet network, an InceptionV3 network, a ResNet network having 50 layers, and a VGG-16 network were trained and benchmarked. The dataset was specifically created for this project, containing 14,478 images distributed across 10 classes A pretrained ImageNet model was fine-tuned and transfer learning was applied. A genetically weighted ensemble of convolutional neural networks achieved a 90% classification accuracy. Arief Koesdwiady et. al.[11] in their study compared two frameworks VGG-19 and XGBoost to classify different driver distraction. They created their own image dataset for this study. Concept of transfer learning was used to train the VGG19 model. The VGG19 model outperformed the XGBoost model by 7%. The dataset presented different challenges in terms of different illumination conditions, camera positions and variations in driver's ethnicity, and genders. The end-to-end framework was able to detect different classes with a best test accuracy of 95% and an average accuracy of 80% per class.

In various other studies distractions and causes of the distractions were detected by using CNNs [20], a combination of CNN and Random Forest for the detection of the distraction categories in images [21], SVM and three end-to-end CNNs, namely, AlexNet, VGG-16 and ResNet-152 [22]. In another study [23] AdaBoost classifier was applied along with hidden Markov models to classify Kinect RGB-D data on staged distractions. A hidden conditional random-fields model was used to detect mobile-phone usage [24]. Similarly, a faster Region Based Convolutional Neural Networks (R-CNN) model was designed [25] to detect phone usage. We also find use of histogram of gradients (HOGs) and AdaBoost classifiers [26] for driver distraction.

On analyzing other computer vision problems it was identified that there are multiple pre-trained models [27] such as ImageNet, Sports 19M, IG-Kinetics 19M Images/250M Images and 19 M Videos. In some cases, short, localized videos performed better than long videos/frames for identifying the actions. Also, faster performance [28] can be obtained by starting with a 2D architecture, and inflating all the filters and pooling kernels – endowing them with an additional temporal dimension. Filters are typically square and we just make them cubic – $N \times N$ filters become $N \times N \times N$. Imagenet is considered as one of the best pre-training models [29], there are multiple factors that are considered in this paper – whether coarse grain/fine grain classes are better. Whether more dataset per class or more classes is better for pre-training. Example taking 50% of pre-trained dataset resulted in drop of just 1.5% but gave a greater performance. Similarly, blindly adding more training data only further impacted performance of processing. The analysis [30] was also done over 14 million imagenet images having over 21k classes to answer questions on right pre-training with appropriate size of classes. In addition to rightly using Imagenet, a very deep convolution network [31] on Imagenet models helped in further scaling of performances. The filter size of 3x3 helped in building deep layers of upto 19 CNNs. Spatial padding of 1 on 3x3 and spatial pooling helps in better performance and action prediction.

III. DATASET DESCRIPTION

The StateFarm distraction-detection dataset is selected for the Capstone project. Dataset was published on Kaggle in 2016 [5] for a competition. This dataset is the most widely used dataset for the detection of driver distraction and has been applied in many studies. The StateFarm dataset includes ten classes as listed below:

- c0: safe driving
- c1: texting – right
- c2: talking on the phone – right
- c3: texting – left
- c4: talking on the phone – left
- c5: operating the radio
- c6: drinking
- c7: reaching behind
- c8: hair and makeup

- c9: talking to passenger

The metadata such as creation dates is removed from the images. State Farm set up these experiments in a controlled environment - a truck dragging the car around on the streets - so these "drivers" weren't really driving. The train and test data are split on the drivers, such that one driver can only appear on either train or test set. The images are a collection of left-hand drive vehicles only. Each class contains close to 2300 images, breakup of images per class is listed below:

TABLE.1. DRIVER ACTIONS AND TOTAL IMAGES IN THE CLASS.

| Classes | Driver actions | Images |
|---|---|---|
| C0 | safe driving | 2489 |
| C1 | texting - right | 2267 |
| C2 | talking on the phone - right | 2317 |
| C3 | texting - left | 2346 |
| C4 | talking on the phone - left | 2326 |
| C5 | operating the radio | 2312 |
| C6 | drinking | 2325 |
| C7 | reaching behind | 2002 |
| C8 | hair and makeup | 1911 |
| C9 | talking to passenger | 2129 |
| | Total | 22424 |

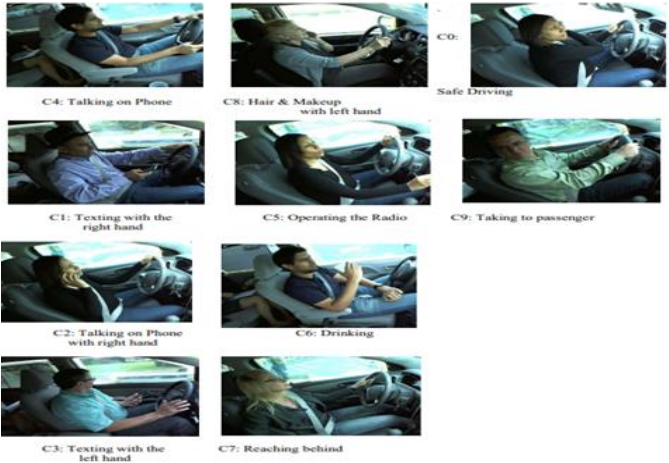

Fig. 1. Image visualization of all 10 classes

### A. Production Dataset -

Additionally, we found that 79726 unlabeled data had distinct drivers and was used to predict the accuracy of the model without allowing data leakage. It was found that these images varied on the drivers, location, position of camera

### B. Evaluation Metrics

We divided the data into train and validation subsets and using the Classification Accuracy of each class to determine the success of the model. A class accuracy above 90% is the set benchmark based on work done in the same field [1][2].

$$Classification\ Accuracy = \frac{No.\ of\ correct\ predictions}{No.\ of\ correct\ predictions}$$

We also checked the performance of the model based on the unlabeled data and the time taken to predict an image. To be able to use the classifier on real-time data the speed of the model to predict and classify the image was also important. Every class is of equal importance to us.

## IV. TECHNICAL APPROACH

Based on the solution diagram mentioned (Fig 2), the intent is to use the pre-trained Imagenet model for weight initialization. Ensemble of CNN models would be implemented to control tuning of the model and making it lightweight and performant for deployment on an edge device. We would also run Resnet50 and VGG16 transfer learning architecture, for building and learning in non-production and comparing with our CNN ensemble.

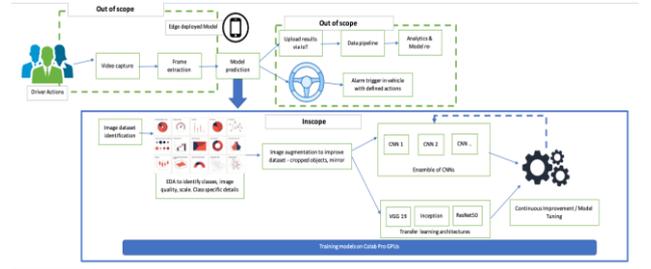

Fig. 2. The proposed ensemble of CNN to predict the classes

### C. Exploratory Data Analysis and visualization

The first step is Exploratory Data Analysis (EDA). It helps us analyze the entire dataset and summaries its main characteristics, like class distribution, size distribution, and so on. The StateFarm distraction-detection dataset is selected for the Capstone project. Dataset was published on Kaggle in 2016 [3] for a competition. This dataset is the most widely used dataset for the detection of driver distraction and has been applied in many studies. The StateFarm dataset includes ten classes.

The train and test data are split on the drivers, such that one driver can only appear on either train or test set. The images are a collection of left-hand drive vehicles only. Each class contains close to 2300 images.

The second step is Image Pre-Processing, where the aim is to take the raw image and improve image data (also known as image features) by suppressing unwanted distortions, resizing and/or enhancing important features, making the data more suited to the model and improving performance.

In our approach to build the model we will visualize the images and do image augmentation. We load the dataset and visualize sample images from each category as shown in Fig 1.

Using Image augmentation to add noise to the model. The model will be able to learn better from noise and have more data to train on. We convert the images to greyscale and resize as seen in Fig 3. We have used the concept of average image to compute and visualize the average image for a specific class (Fig 4) and then compute the difference between the averages to visualize the contrast as seen in Fig 5

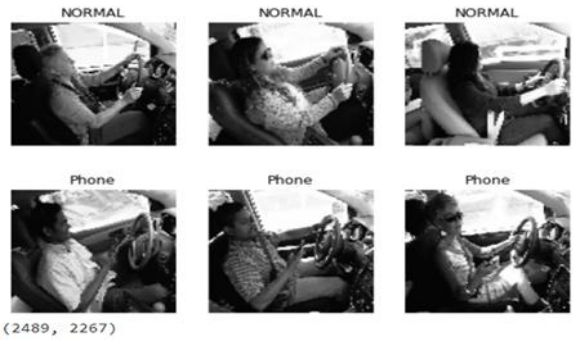

Fig. 3.  Greyscale images*

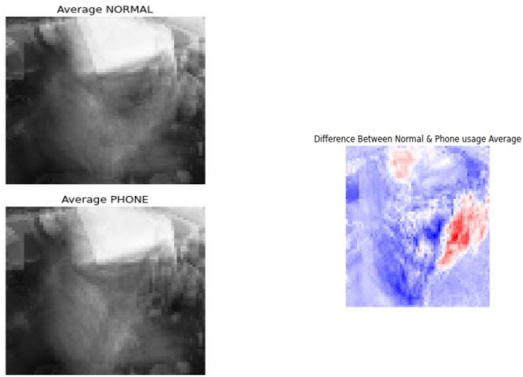

Fig. 4. Calculated the average image for said class*    Fig. 5. Computed the difference between the average of the classes*

* Comparison between c0 and c1 category. c0 category is designated as Normal and c1 category is designated as Phone in the above images

It is noteworthy that due to specific nature of problem, image augmentation on test data (rotation, blurring etc.) results in unnecessary noise or making the images useless which does not support training the model.

### D. Final CNN Model

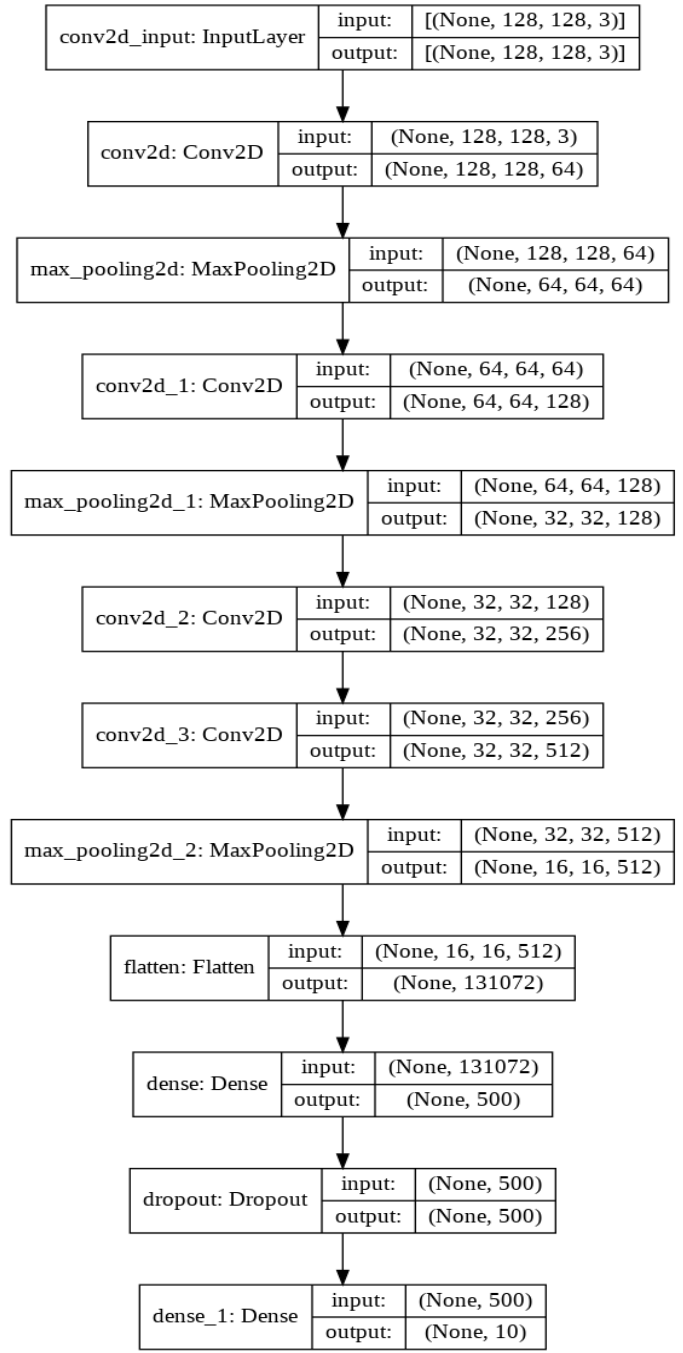

## E. Final VGG-16 Architecture

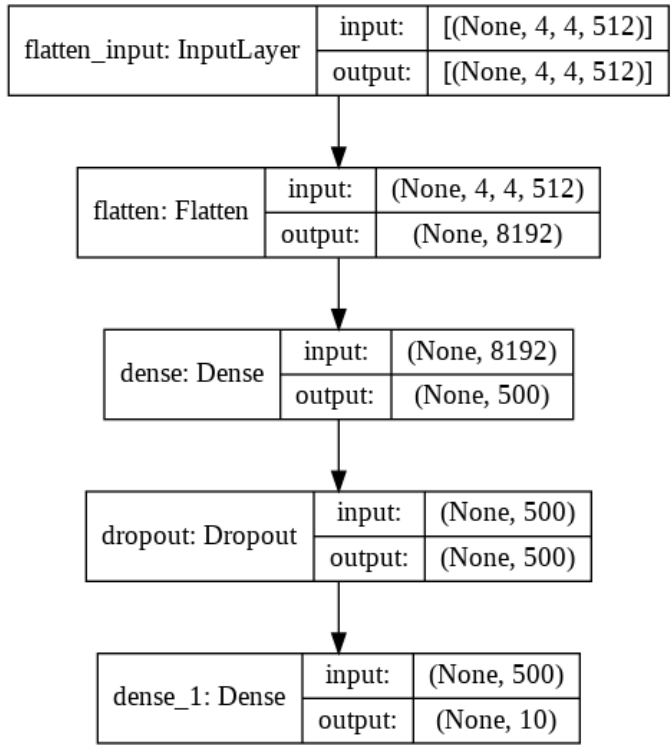

## F. Key Parameters

TABLE 2. KEY PARAMETERS OF EACH MODEL

| Model | Epochs | Batch size | Dense Layers | Conv2D Layers | Pool Layers | Dropouts | Activation function | Kernel Optimization | Trainable Parameters | Learning Rate |
|---|---|---|---|---|---|---|---|---|---|---|
| CNN | 25 | 40 | 2 | 4 | 4 size (2,2) | 2 (0.5) | softmax | glorot_normal | 17,079,366 | 0.001 |
| CNN Optimized | 25 | 40 | 2 | 4 | 3, size (2,2) | 1 (0.5) | softmax | He_normal | 67,092,486 | 0.001 |
| VGG16 | 400 | 16 | N/A | N/A | N/A | N/A | softmax | glorot_normal | 5,130 | 0.001 |
| VGG16 Optimized | 75 | 16 | 2 | N/A | N/A | 1 (0.5) | softmax | glorot_normal | 4,101,510 | 0.001 |
| Resnet | 40 | 16 | N/A | N/A | N/A | N/A | softmax | - | 20,490 | 0.001 |

## V. RESULTS AND DISCUSSION

F1 score for all models and the time taken for prediction robust solution

Table 2. Accuracy of individual model

| Model | Accuracy Scores | |
|---|---|---|
| | Validation Dataset | Production/Test Dataset |
| **CNN** | 99.465 | 67.76 |
| **CNN Optimized** | 99.554 | 69.42 |
| **VGG16** | 96.276 | 52.89 |
| **VGG16 Optimized** | 99.33 | 62.81 |
| **Resnet50** | 89.81 | 33.05 |

TABLE 3. F1 SCORE OF ALL MODELS ON PRODUCTION DATASET

| Model | F1 SCORE | PREDICTION TIME |
|---|---|---|
| CNN | 67.76 | <1 s |
| CNN Optimized | 69.42 | <1 s |
| VGG16 | 52.89 | <1 s |
| VGG16 Optimized | 62.68 | <1 s |
| Resnet | 33.59 | <1 s |
| CNN Ensemble | 0.6630 | <1 s |
| CNN & VGG 16 Ensemble | 0.7240 | <1 s |

TABLE 4. ACCURACY FOR EACH CLASS VS MODEL

| Classes | CNN | CNN Optimized | VGG16 | VGG16 Optimized | Resnet | CNN Ensemble | CNN & VGG 16 Ensemble |
|---|---|---|---|---|---|---|---|
| C0 | 78 | 65 | 56 | 39 | 21 | 56 | 52 |
| C1 | 53 | 46 | 53 | 61 | 23 | 53 | 61 |
| C2 | 54 | 72 | 36 | 36 | 27 | 54 | 72 |
| C3 | 83 | 75 | 16 | 58 | 16 | 66 | 83 |
| C4 | 60 | 70 | 40 | 70 | 50 | 70 | 70 |
| C5 | 100 | 100 | 64 | 100 | 57 | 100 | 100 |
| C6 | 66 | 93 | 80 | 100 | 26 | 93 | 93 |
| C7 | 58 | 58 | 56 | 58 | 33 | 50 | 66 |
| C8 | 50 | 50 | 25 | 50 | 25 | 75 | 100 |
| C9 | 28 | 28 | 57 | 42 | 71 | 28 | 28 |

VGG 16 is not completely optimized as we found opportunity for adding additional layers. Some of the classes have duplicate images which prevent in accurate training for those classes and results in sampling /training errors. Classes C9 and C0 are very similar and all models had problem in predicting the classes C0 & C9 correctly. All models were easily able to predict class C5 (100% prediction) and in some C8 class as well

## G. Compariosn to Benchmark

A validation accuracy above 90% is the set benchmark based on work done in the same field [7][10]. Our models performed way better achieving an accuracy of up to 99.5% with modified CNN model.

## H. Visualization

We have included the visualization for CNN modified model. Please find the visualization for all the other models in the appendix:

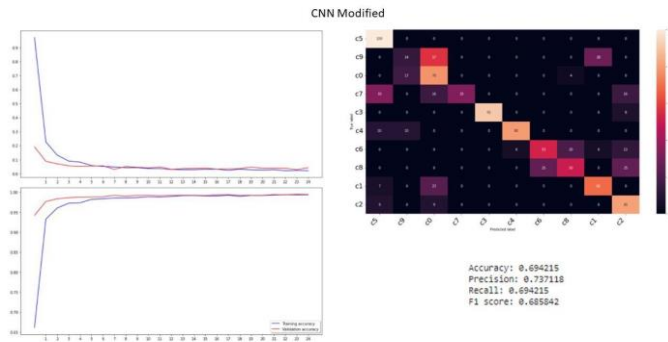

## VI. CONCLUSION AND FUTURE WORK

The fine tuning of the model and optimization of parameters is a step, in the direction for being able to better and faster synthesize the driver behavior.

## I. Limitation

Limitations and closing reflections go hand in hand. State of the art models.

- The labelled dataset available gave an excellent accuracy for each class but on production data testing the model was only ~70% accurate, this is because of data leakage which resulted in good train and val accuracy.
- Large volume and size of images resulted in colab crashing several times, creating multiple models required several accounts and parallel processing.
- Understanding of deployment of model on 5G devices and executing in real time is limited in team, hence further adaptation and optimization of the solution was only limited

## J. Closing Reflection

- Due to dynamic nature of the problem a highly diverse dataset needs to be defined which helps in tuning the models across all actions and frame of references.
- Leveraging highly scalable and compute intensive cloud infrastructure such as Google cloud or Azure cloud for handling transfer learning model on the dataset of over 4 GB fixed bandwidth solution.
- Improve learning of the transfer learning model and the impact of multiple layers as it was found after executing multiple architectures that creating additional layers did not improve the accuracy.

## APPENDIX

Model results and visualizations for the models:

### A. CNN

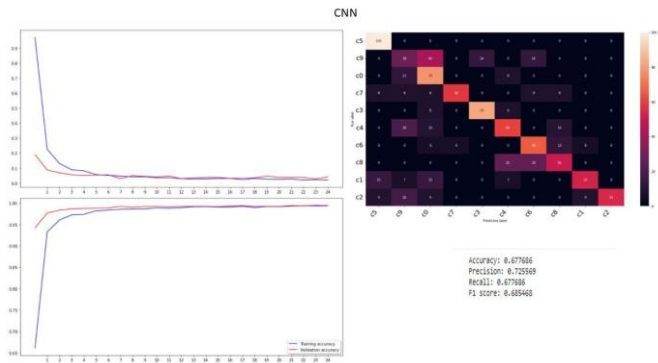

### B. VGG16

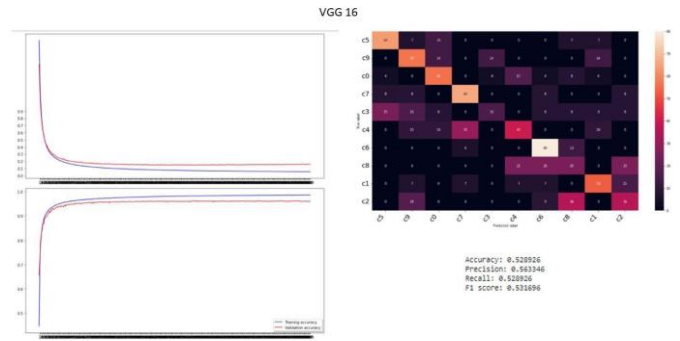

### C. VGG16 Modified

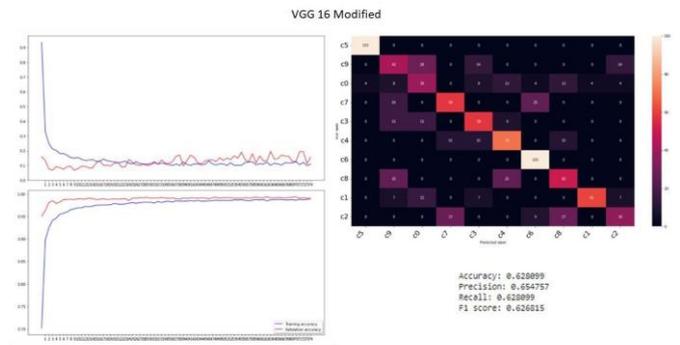

### D. RestNet50

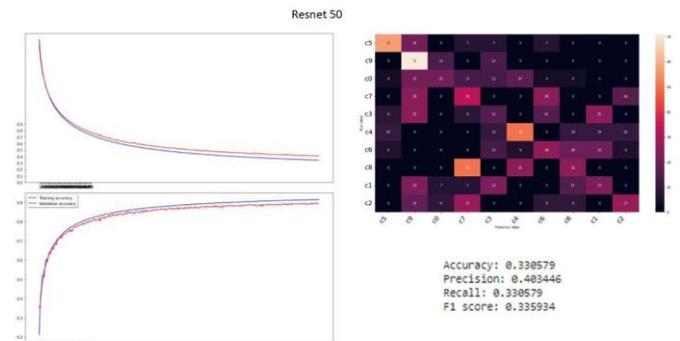